\documentclass[conference]{IEEEtran}
\IEEEoverridecommandlockouts
\usepackage{cite}
\usepackage{amsmath,amssymb,amsfonts}
\usepackage{graphicx}
\usepackage{textcomp}
\usepackage{xcolor}
\usepackage{booktabs}
\usepackage{multirow}
\usepackage{xspace}
\usepackage{url}
\usepackage{tikz}
\usetikzlibrary{arrows.meta,positioning,fit,calc,shapes.arrows}
\usepackage{pgfplots}
\pgfplotsset{compat=1.17}
\usepgfplotslibrary{groupplots}

\newcommand{\method}{ToRA\xspace}
\newcommand{\methodlong}{Tandem of low-Rank Adapters\xspace}

\definecolor{torablue}{HTML}{2A78D6}
\definecolor{qgorange}{HTML}{EB6834}
\definecolor{spaqua}{HTML}{1BAF7A}

\newcommand{\std}[1]{\textsubscript{\,$\pm$#1}}
\newcommand{\best}[1]{\textbf{#1}}
\newcommand{\parhead}[1]{\noindent\textit{#1.}\ }
\newcommand{\us}{\,$\mu$s\xspace}

\begin{document}

\title{Run One, Get Two: Near-Free Streaming Safety Checks for Local LLMs}

\newif\ifblind
\blindfalse
\ifblind
\author{\IEEEauthorblockN{Anonymous Authors}
\IEEEauthorblockA{\textit{Paper under double-blind review}}}
\else
\author{\IEEEauthorblockN{Soheil Zibakhsh\IEEEauthorrefmark{1}\thanks{\IEEEauthorrefmark{1}Corresponding author: \texttt{szibakhsh@ucsd.edu}}, Pedram Aghazadeh, and Farinaz Koushanfar}
\IEEEauthorblockA{\textit{University of California San Diego}, La Jolla, CA, USA}}
\fi

\maketitle

\begin{abstract}
Local LLM assistants decode for one user at a time, so decoding is bound by memory bandwidth and leaves compute idle. Streaming safety checks today trade accuracy against cost. A dedicated guard model is accurate but nearly doubles the latency per token, while a hidden-state probe is cheap but less accurate. We present \method, an adapter that runs a second model as an extra row in the generator's decode step, sharing its weights and KV cache. Dual-row kernels read each weight once for both rows. On Qwen3-0.6B, \method matches or beats a dedicated streaming guard, with higher F1 on 12 of 17 benchmarks. It detects 11.8 points more unsafe responses than the probe at a similar cost, under 4\% added latency per token.


\end{abstract}


\section{Introduction}



Local LLM assistants must perform generation and safety monitoring on the same device, under tight latency and memory constraints. Checking only completed responses cannot prevent unsafe content from reaching users. Streaming safety checkers instead evaluate tokens as they are generated, allowing unsafe responses to be interrupted early~\cite{qwen3guard,scm}. The challenge is to provide accurate detection without substantially slowing decoding.

\begin{figure}[t]
\centering
\begin{tikzpicture}[
  gain/.style={single arrow, draw=none, fill=black, fill opacity=0.07, text opacity=1, text=black!60,
    font=\scriptsize, inner sep=1.5pt, single arrow head extend=2.5pt, single arrow head indent=0pt},
]
\begin{axis}[
  scale only axis, width=6.6cm, height=2.9cm,
  axis lines=left, axis line style={black!55, -}, tick style={black!55}, tick align=outside,
  xmin=-4, xmax=100, xtick={0,25,50,75,100}, xticklabels={0\%,25\%,50\%,75\%,100\%},
  ymin=76.9, ymax=81.9, ytick={77,78,...,81},
  xlabel={Added latency per token}, ylabel={Avg.\ response F1},
  tick label style={font=\scriptsize}, label style={font=\footnotesize},
  xlabel style={yshift=2pt}, ylabel style={yshift=-3pt},
  ymajorgrids, grid style={black!7, line width=0.3pt},
  clip=false,
]
  \addplot[only marks, mark=square*, mark size=2.2pt, color=spaqua] coordinates {(1.63,77.6)};
  \addplot[only marks, mark=triangle*, mark size=2.8pt, color=qgorange] coordinates {(85.86,80.4)};
  \addplot[only marks, mark=*, mark size=2.4pt, color=torablue, mark options={draw=white, line width=0.4pt},
    error bars/.cd, y dir=both, y explicit, error bar style={line width=0.6pt}] coordinates {(3.52,81.4) +- (0,0.2)};
  \node[anchor=west, font=\scriptsize, text=spaqua!80!black] at (axis cs:4.2,77.6) {SingProbe (+1.6\%)};
  \node[anchor=west, font=\scriptsize\bfseries, text=torablue] at (axis cs:6.0,81.4) {\method{} (+3.5\%)};
  \node[anchor=north east, font=\scriptsize, text=qgorange!85!black] at (axis cs:100,80.08) {Qwen3Guard-Stream (+85.9\%)};
  \coordinate (effTail) at (axis cs:82.0,80.4);
  \coordinate (effTip)  at (axis cs:10.0,80.4);
  \coordinate (accTail) at (axis cs:2.6,77.9);
  \coordinate (accTip)  at (axis cs:2.6,81.05);
\end{axis}
\path let \p1=($(effTip)-(effTail)$), \n1={veclen(\x1,\y1)} in
  node[gain, minimum height=\n1, shape border rotate=180, anchor=tail] at (effTail) {$24\times$ less overhead};
\path let \p1=($(accTip)-(accTail)$), \n1={veclen(\x1,\y1)} in
  node[gain, minimum height=\n1, rotate=90, anchor=tail] at (accTail) {$+3.8$ F1};
\end{tikzpicture}
\caption{Response F1 versus per-token latency overhead at 2k context. All methods use identical generator kernels; the error bar shows one standard deviation across three \method seeds.}
\label{fig:teaser}
\end{figure}
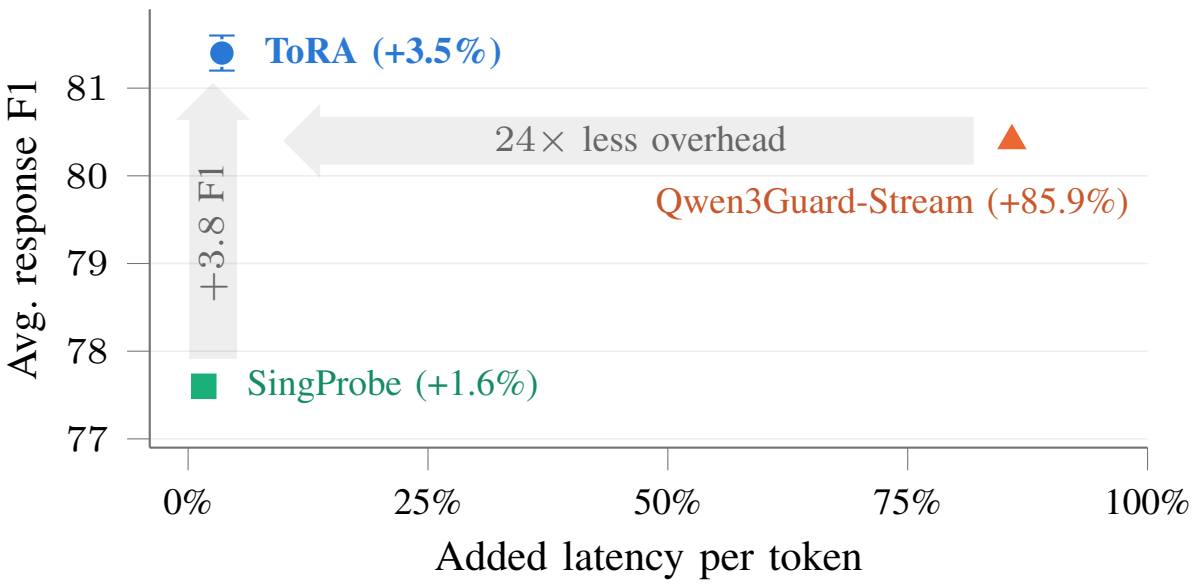

\begin{figure}[t]
\centering
\includegraphics[width=\columnwidth, trim=86 17 20 14.5, clip]{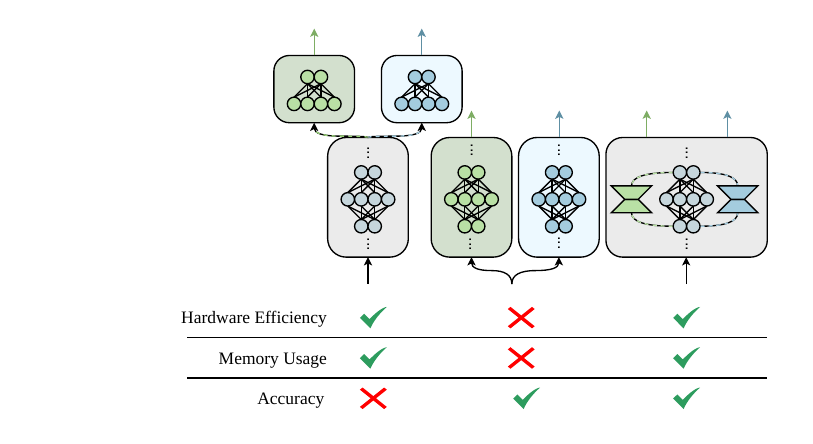}
\caption{Streaming safety architectures: hidden-state probe (left), dedicated guard (center), and \method (right).}
\label{fig:compare}
\end{figure}

Existing streaming checkers face a fundamental trade-off between accuracy and efficiency (Figs.~\ref{fig:teaser} and~\ref{fig:compare}). Dedicated guard models, such as Qwen3Guard-Stream~\cite{qwen3guard}, provide strong detection through a separately trained language model. However, running an additional model requires its own weights and key-value (KV) cache. On an A6000 GPU, pairing Qwen3Guard-Stream with a generator of the same size increases per-token latency by 84--89\% and consumes approximately 2~GiB of additional memory at an 8k-token context length. Lightweight alternatives, such as
SingProbe~\cite{singprobe}, instead classify the generator's hidden states. This avoids the cost of a separate model and adds less than 2\% latency. However, the underlying representations are not explicitly optimized for safety detection. Consequently, SingProbe trails Qwen3Guard-Stream by 2.8 points in response F1 and detects nearly 13\% fewer unsafe responses during streaming.

At batch size one, auto-regressive decoding is typically memory-bandwidth bound. Each step loads the model weights but performs relatively little computation, approximately one floating-point operation per byte in our setting, far below the compute-bound regime~\cite{roofline}.
This unused capacity could support verification at minimal cost if both paths share weights and avoid additional memory traffic. However, conventional LoRA execution requires a separate KV cache or repeated weight reads. Even with cache sharing, standard execution adds
15--52\% decoding latency in our experiments (Section~\ref{sec:abl}). Efficient verification therefore requires shared weight accesses and cached representations without altering the generator's output.



We introduce \method (\methodlong), a shared-backbone architecture that performs generation and safety verification within a single decoding step. Rather than running an independent guard model, \method adapts the generator's upper $K$ transformer layers using low-rank adapters (LoRA)~\cite{lora} and a lightweight classification head. The verifier path branches from the generator at the first adapted feed-forward layer, reusing all preceding computations. It also utilizes the generator's cached keys and values from earlier tokens, eliminating the need for a separate KV cache. To execute both paths efficiently, we develop dual-row Triton~\cite{triton} kernels that load each shared weight tile once and apply the low-rank updates only to the verifier path. The generator's output remains unaffected by verification.

Fig.~\ref{fig:compare} compares the three streaming safety architectures, while Fig.~\ref{fig:teaser} illustrates their accuracy--latency trade-off. On Qwen3-0.6B~\cite{qwen3} \method improves average response and prompt F1 over Qwen3Guard-Stream by 1.0 and 1.7 points, respectively, with higher F1 on 12 of 17 benchmarks. Its streaming detection remains within 1.5 points of the guard,
while identifying unsafe content earlier. Compared with SingProbe, \method improves response F1 by 3.8 points and detects 11.8 percentage points more unsafe responses at a 5\% false-alarm budget. At 2k context, it adds only 3.5\% latency per token, compared with 1.6\% for SingProbe and 85.9\% for Qwen3Guard-Stream. The additional 60~\us per token over SingProbe amounts to 30~ms over 500 tokens. \method also requires only 9.3~MiB of additional memory and no separate KV cache.

Our contributions are fourfold: \textit{(i)} a shared-backbone verifier that preserves the generator's output; \textit{(ii)} efficient token-level training using one frozen generator pass per sequence; \textit{(iii)} dual-row Triton kernels that reuse weight accesses across generation and verification; and \textit{(iv)} evaluation across 17 safety benchmarks and streaming detection, demonstrating
guard-level accuracy at nearly the cost of a probe. \ifblind We will release our implementation and kernels upon acceptance.\else Our code is available at \url{https://github.com/soheilzi/ToRA}.\fi

\section{Background and Related Work}
\label{sec:background}

Conventional guards, including Llama Guard~\cite{llamaguard}, WildGuard~\cite{wildguard}, ShieldGemma~\cite{shieldgemma}, Aegis~\cite{aegis1,aegis2}, and Qwen3Guard-Gen~\cite{qwen3guard}, classify prompts or completed responses. Streaming checkers
instead monitor partial outputs during generation. Qwen3Guard-Stream~\cite{qwen3guard} uses token-level classification heads but requires separate weights and a KV cache. SingProbe~\cite{singprobe} avoids this overhead by probing hidden states from three generator layers,
although these representations are not optimized for safety detection. \method combines task-specific adaptation with shared weights and cache.

\parhead{Adapters and multi-adapter serving} LoRA~\cite{lora} adapts frozen models through low-rank weight updates. S-LoRA~\cite{slora} and Punica~\cite{punica} batch requests with different adapters and independent KV caches. Activated LoRA~\cite{alora} reuses the base model's prefix cache but does not perform continuous concurrent execution. LoRA-Guard~\cite{loraguard} switches a language model between generation and safety classification. Unlike these methods, \method
performs token-level verification using shared weights, a shared cache, and fused GPU kernels.


\parhead{Compute utilization during decoding} Speculative decoding~\cite{specdec}, Medusa~\cite{medusa}, and EAGLE~\cite{eagle} exploit parallel token verification to accelerate generation. Tandem Transformers~\cite{tandem} coordinate models of different sizes, while RoE~\cite{roe} aggregates parallel expert outputs. CALM~\cite{calm} and LayerSkip~\cite{layerskip} instead reduce computation through early exits. Unlike these generation-oriented
approaches, \method uses available compute for continuous safety verification.


\section{Method}
\label{sec:method}
\begin{figure}[t]
\centering
\begin{tikzpicture}[
  font=\footnotesize, >={Stealth[length=4pt]}, line width=0.5pt,
  blk/.style={draw=black!55, rounded corners=1.5pt, minimum width=2.45cm, minimum height=0.46cm, align=center, inner sep=1.5pt},
  gen/.style={blk, fill=black!7},
  ver/.style={blk, draw=torablue, fill=white},
  ada/.style={blk, draw=torablue, fill=torablue!18},
  note/.style={align=center, text=black!60, font=\scriptsize\itshape},
  ]
  \def\xg{1.3} \def\xc{4.1} \def\xv{7.1}
  \node (tok) at (\xg,0) {token $x_t$};
  \node[gen] (emb) at (\xg,0.6) {Embedding};
  \node[gen] (low) at (\xg,1.28) {Layers $0,\dots,L{-}K{-}1$};
  \node[gen] (attk) at (\xg,1.96) {Attention, layer $L{-}K$};
  \node[gen] (ffnk) at (\xg,2.64) {FFN $W$};
  \node[gen] (att) at (\xg,3.6) {Attention};
  \node[gen] (ffn) at (\xg,4.36) {FFN $W$};
  \node[gen] (lm) at (\xg,5.42) {Norm, LM head};
  \node (out) at (\xg,6.02) {next token};
  \node[note] at (\xv,1.28) {not computed: $h^V = h^G$};
  \node[ada] (ffnkv) at (\xv,2.64) {FFN $W + sBA$};
  \node[ver] (attv) at (\xv,3.6) {Attention};
  \node[ada] (ffnv) at (\xv,4.36) {FFN $W + sBA$};
  \node[ada] (head) at (\xv,5.42) {Norm, head};
  \node (score) at (\xv,6.02) {score $p_t$; stop if $p_t \ge \tau$};
  \node[draw=black!55, fill=black!3, minimum height=0.46cm, align=center, inner sep=2pt] (kv) at (\xc,3.6) {$G$'s KV cache};
  \foreach \a/\b in {tok/emb, emb/low, low/attk, attk/ffnk, ffnk/att, att/ffn, ffn/lm, lm/out} \draw[->] (\a) -- (\b);
  \draw[->, torablue] (attk.east) -| (ffnkv.south);
  \foreach \a/\b in {ffnkv/attv, attv/ffnv, ffnv/head, head/score} \draw[->, torablue] (\a) -- (\b);
  \draw[->] (att.east) -- node[above=-1pt, font=\scriptsize] {write} (kv.west);
  \draw[->, dashed, torablue] (kv.east) -- node[below=-1pt, font=\scriptsize] {read ${<}t$} (attv.west);
  \draw[black!45, densely dotted] (ffnk.east) -- node[below=-1pt, note] {one read of $W$} (ffnkv.west);
  \draw[black!45, densely dotted] (ffn.east) -- node[below=-1pt, note] {one read of $W$} (ffnv.west);
  \node[draw=black!40, dashed, rounded corners=2pt, fit=(att)(ffn)(attv)(ffnv), inner sep=3pt] (rep) {};
  \node[note, anchor=south] at (\xc,4.7) {repeated for layers $L{-}K{+}1,\dots,L{-}1$};
  \node[font=\footnotesize\bfseries] at (\xg,6.55) {Generator row};
  \node[font=\footnotesize\bfseries, text=torablue] at (\xv,6.55) {Verifier row};
\end{tikzpicture}
\caption{\method decoding step. The verifier branches at the first adapted FFN and reuses the generator's weights and KV cache. Dotted lines indicate shared weight reads.}
\label{fig:method}
\end{figure}
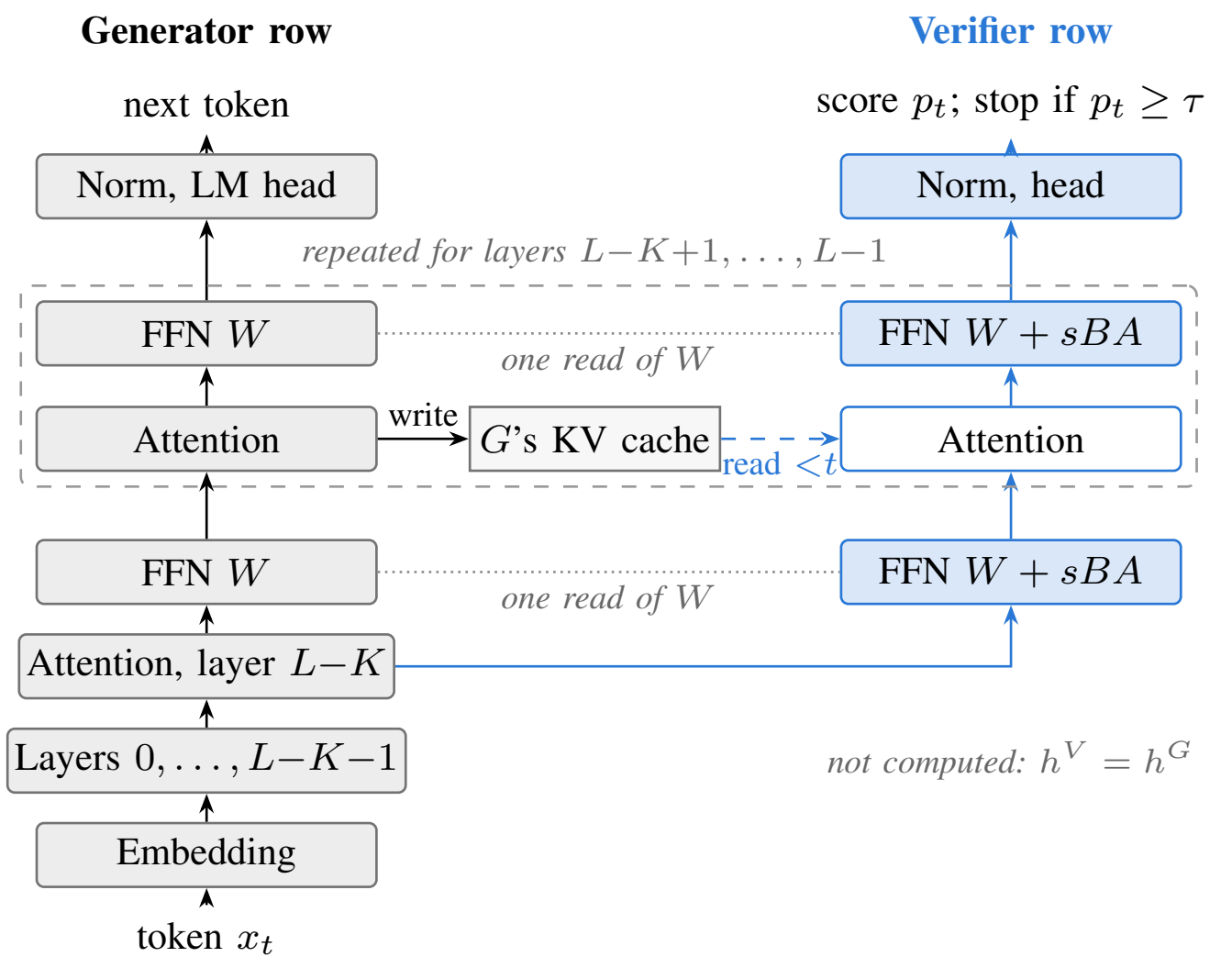

\method performs generation and safety verification within a single decoding step. As illustrated in Fig.~\ref{fig:method}, the verifier reuses the generator's weights and cached representations, while low-rank adapters specialize its upper layers for safety classification. The two paths execute together using shared weight accesses, avoiding the overhead of a separate guard model.

\subsection{Verifier Architecture}

Consider a decoder-only transformer generator $G$ with $L$ layers, each consisting of an attention block followed by a feed-forward network (FFN). Each block has its own input normalization. We construct a verifier $V$ by adapting the top $K$ layers of $G$, while keeping all original model weights frozen.

Specifically, we introduce low-rank updates to the gate, up, and down projections of each adapted FFN. For a projection matrix $W$, the verifier computes \begin{equation}
\label{eq:lora}
y = Wx + s\,B(Ax), \qquad
A \in \mathbb{R}^{r\times d_{\mathrm{in}}},\;
B \in \mathbb{R}^{d_{\mathrm{out}}\times r},
\end{equation}
where $r$ denotes the adapter rank and $s$ is its scaling factor. A lightweight classification head replaces the language-model head, mapping the verifier's final hidden state to $C$ output scores. Thus, $V$ predicts safety labels rather than generating tokens.



At decoding step $t$, the generator executes its lower $L-K$ layers once. The verifier branches after attention in layer $L-K$, immediately before the first adapted FFN. Above this point, the generator applies the original projections, while the verifier uses Eq.~\eqref{eq:lora}. Both retain the original attention weights, and their FFN computations are fused as described in Section~\ref{sec:kernels}.

\subsection{Shared KV Cache}

A conventional adapted model maintains its own KV cache to store the keys and values of previously processed tokens. This introduces additional memory usage and traffic that grow quadratically with the sequence length. \method avoids this overhead by allowing the verifier to reuse the generator's cache.

At each attention layer following the branching point, the verifier computes its own query, key, and value for the current token. However, instead of maintaining historical keys and values, it attends to those already stored by the generator. Formally, the verifier's attention
output at position $t$ is

\begin{equation}
\label{eq:shared_attn}
o^V_t = \mathrm{softmax}\!\left(\frac{q^V_t\,\bigl[K^G_{<t};\,k^V_t\bigr]^{\top}}{\sqrt{d}}\right)\bigl[V^G_{<t};\,v^V_t\bigr].
\end{equation}

where $K^G_{<t}$ and $V^G_{<t}$ are the generator's cached keys and values from earlier positions. The verifier computes $k^V_t$ and $v^V_t$ only for the current token and discards them after attention. Consequently, \textbf{only the generator updates the persistent KV cache}, and the verifier requires no additional cache that grows with the context length.

This design allows the verifier to process previous tokens through the generator's representations rather than recomputing its own. The generator, in turn, never accesses the verifier's states, preserving the original generation computation. We use the same cache-sharing mechanism during training and inference, and evaluate its effect on classification accuracy in Section~\ref{sec:abl}.

Although we focus on a single safety verifier, the architecture is not restricted to this task. Additional adapters could support other verification tasks while reusing the same generator weights and cached states.




\subsection{Efficient Training}
\label{sec:training}

We train the verifier on fixed prompt--response sequences with the generator frozen. A single forward pass of $G$ provides the hidden states at the branching point and the KV states of the adapted layers. The verifier then processes only the supervised positions, attending to the corresponding generator KV states as in Eq.~\eqref{eq:shared_attn}. This reproduces the decoding
computation without a separate verifier cache. Only the adapters and classification head are updated using class-weighted binary cross-entropy.

\subsection{Token-Level Supervision}


Labels assigned only to the final tokens of prompts and responses are sufficient for classifying complete inputs. However, they provide no direct supervision for detecting unsafe content as it emerges during generation. A verifier trained exclusively on final-token labels may therefore classify completed responses accurately while failing to detect unsafe content early.

To address this limitation, we introduce supervision at $N$ additional positions within each training response, sampled uniformly without replacement. For safe responses, all sampled positions receive a label of 0. For unsafe responses, a token-level teacher identifies the onset of unsafe content. Positions preceding this onset receive a label of 0, while positions at or after the onset receive a label of 1.

When the teacher cannot identify an unsafe onset, the sampled positions are labeled 0, while the final response token retains its original unsafe label. The intermediate labels collectively receive the same loss weight as the final response label, balancing token-level supervision with response-level classification. Importantly, the teacher is applied only to responses already labeled unsafe, leaving the labels of safe responses unaffected.

During inference, the verifier evaluates the response safety score at each generated token. If the score reaches a predefined threshold $\tau$, the response is flagged and generation can be stopped immediately. This enables continuous safety monitoring without waiting for the complete response.

\section{Kernel and System Design}
\label{sec:kernels}

To minimize verification overhead, \method must exploit the unused compute capacity of memory-bound decoding while preserving the generator's output. Conventional execution approaches introduce unnecessary memory traffic or computation. Executing the generator and verifier separately requires reading the shared weights twice. Similarly, merging the adapters into a separate copy of the weights, $W+sBA$, duplicates weight storage and reads. Batching both paths into a single cuBLAS GEMM avoids redundant weight reads, but leaves the adapter computations as separate operations. Moreover, changing the number of input rows can alter the GEMM reduction order, leading to numerical differences in the generator's logits~\cite{batchinv}. To address these limitations, we develop two Triton kernels that execute both paths with shared weight accesses while preserving independent computations for each row.


\parhead{Dual-row linear projection}
Our \texttt{dual\_lora\_linear} kernel computes the generator and verifier projections simultaneously: $y_0=Wx_0$ and $y_1=Wx_1+sB(Ax_1)$, respectively. Each Triton program loads a tile of the shared weight matrix $W$ once and updates separate FP32 accumulators for the two rows. The low-rank projection $Ax_1$ is accumulated within the same loop. The kernel then applies the adapter contribution $sB(Ax_1)$ to the verifier's output in its epilogue, leaving the generator's result unmodified. The additional adapter weight reads amount to $r(d_{\mathrm{in}}+d_{\mathrm{out}})$ elements, compared with $d_{\mathrm{in}}d_{\mathrm{out}}$ elements for the shared projection. At the rank used in our experiments (Section~\ref{sec:setup}), this represents approximately 1\% of the base weight traffic.


\parhead{Fused FFN execution}
To further reduce kernel launches, we introduce \texttt{dual\_lora\_gate\_up}, which fuses the gate and up projections of the FFN. The kernel computes both projections for the generator and verifier in a shared weight traversal, applying the low-rank updates only to the verifier. It then evaluates the activation $\mathrm{SiLU}(g)\odot u$ within the same kernel. The epilogue also produces partial sums of $A_{\mathrm{down}}h_1$, which are consumed by the subsequent down projection. Together with \texttt{dual\_lora\_linear}, this reduces the
entire FFN to two kernel launches for both paths. In comparison, a conventional cuBLAS implementation requires three GEMMs and a separate activation kernel for the generator alone.



\parhead{Preserving generator outputs}
An essential requirement of \method is that verification must not affect the generator's predictions. However, GPU kernel configurations and floating-point reduction orders can vary with the number of input rows. We therefore select kernel configurations based solely
on weight dimensions and perform auto-tuning with two rows, regardless of whether verification is enabled. When the verifier is disabled, FP32 placeholders replace the adapter pointers, preserving the generator's execution path and reduction order.

We verify bit-wise equivalence of the generator row with and without verification across all tested kernel configurations. The complete decoding path also produces identical logits without compilation. With \texttt{torch.compile}, differences remain within the floating-point rounding introduced by compilation itself.

\parhead{Attention and decode execution}
We implement attention using the decoding path of FlexAttention~\cite{flexattn} with a static block mask. The verifier introduces an additional query row that accesses the generator's KV cache, allowing both paths to reuse the same cached keys and values within a single attention operation. We explicitly pack grouped-query attention heads into rows to support this execution pattern.

Only the generator executes layers below the branching point $L-K$, using standard cuBLAS operations. The remaining layers use our dual-row
kernels. To minimize runtime overhead, we maintain static tensor shapes, a single KV-cache tensor per layer, and in-place cache updates. This enables compilation with \texttt{torch.compile}~\cite{pytorch2} and CUDA graph capture, with the Triton kernels integrated as opaque custom operators.


\begin{table*}[t]
\centering
\caption{Accuracy, streaming detection and cost of the three checkers on Qwen3-0.6B.}
\label{tab:main}
\footnotesize
\setlength{\tabcolsep}{3.1pt}
\begin{tabular}{@{}lcccccccccc@{}}
\toprule
 & \multicolumn{2}{c}{Avg. F1} & \multicolumn{4}{c}{Streaming detection} & \multicolumn{3}{c}{Latency per token} &
 Added memory \\
\cmidrule(lr){2-3}\cmidrule(lr){4-7}\cmidrule(lr){8-10}
Method & Response (8) & Prompt (9) & Det.\ @5\% & Exact @5\% & Delay @5\% & Det.\ @1\% & ms at 2k &
\multicolumn{2}{c}{Added at 512 / 2k / 8k} & at 8k \\
\midrule
Qwen3Guard-Stream & 80.4 & 84.6 & \best{84.7} & \best{77.1} & 7 & \best{61.0} & 5.527 &
\multicolumn{2}{c}{+84.3 / +85.9 / +89.3\%} & 1,139 + 910 MiB \\
SingProbe & 77.6 & 83.0 & 71.8 & 61.5 & 7 & 48.5 & \best{3.022} &
\multicolumn{2}{c}{\best{+1.7 / +1.6 / +1.2\%}} & \best{4.3 + 2.0 MiB} \\
\method (ours) & \best{81.4}\std{0.2} & \best{86.3}\std{0.4} & 83.6\std{0.6} & 75.6\std{0.1} & \best{5.0} &
59.5\std{1.6} & 3.078 & \multicolumn{2}{c}{+4.0 / +3.5 / +2.8\%} & 9.3 + 0 MiB \\
\bottomrule
\end{tabular}
\end{table*}
\section{Experimental Setup}
\label{sec:setup}

\parhead{Model configuration}
We use Qwen3-0.6B~\cite{qwen3} with 28 layers and a hidden size of 1,024, in bf16 with thinking mode disabled. The verifier adapts the gate, up, and down projections of the top $K=14$ layers using rank $r=8$ and $\alpha=16$. Its classification head is an MLP with a 1,024-unit hidden layer and dropout 0.1. The two outputs predict prompt and response harm, evaluated at the final prompt and response tokens, respectively. The verifier contains \textbf{2.43M} trainable parameters.



\parhead{Training data}
We use the official training splits of seven datasets: WildGuardMix~\cite{wildguard}, Aegis 2.0~\cite{aegis2}, BeaverTails (30k)~\cite{beavertails}, PKU-SafeRLHF (v0, one response pair per prompt)~\cite{saferlhf}, ToxicChat (0124)~\cite{toxicchat}, Aegis 1.0~\cite{aegis1}, and ExpGuardTrain~\cite{expguard}. We exclude training prompts appearing in any evaluation or validation split, then reserve 5\% of each source for prompt-disjoint validation. The resulting mixture contains 33.6M tokens, 109,883 prompt labels, and
102,072 response labels.


\parhead{Token-level supervision}
We sample $N=128$ intermediate positions per response and use Qwen3Guard-Stream-0.6B~\cite{qwen3guard} to locate unsafe-content onsets. An onset is the first token in the earliest pair of consecutive tokens labeled unsafe or controversial. The teacher identifies an onset in
88.4\% of unsafe training responses and is not applied to safe responses.


\parhead{Optimization}
We train for two epochs using AdamW, a learning rate of $10^{-3}$, zero weight decay, and gradient clipping at norm 1.0. Batches contain 32,768 tokens, with a maximum sequence length of 2,048. The learning rate follows a 3\% linear warm-up and cosine decay to zero. We evaluate four checkpoints and select the one with the highest mean validation AUPRC. Each experiment uses three random seeds.


\parhead{Benchmarks}
We evaluate on the public benchmarks reported by Qwen3Guard and SingProbe. The eight response benchmarks are WildGuardTest, Aegis 2.0, BeaverTails, PKU-SafeRLHF, HarmBench~\cite{harmbench}, Think (Qwen3GuardTest reasoning traces~\cite{qwen3guard}), XSTest-Response~\cite{wildguard}, and ExpGuardTest. The nine prompt benchmarks are WildGuardTest, Aegis 2.0, ToxicChat, OpenAI
Moderation~\cite{openaimod}, Aegis 1.0, SimpleSafetyTests~\cite{sst}, HarmBench, XSTest~\cite{xstest}, and ExpGuardTest. For streaming evaluation, we use Qwen3GuardTest's \texttt{response\_loc} split, which identifies the first unsafe sentence in 813 responses, and SingStreamBench (210 responses)~\cite{singprobe}.


\parhead{Metrics}
We report unsafe-class F1, averaged equally across benchmarks. For \method, we select one threshold per task by maximizing F1 on pooled WildGuard and Aegis 2.0 validation data. Streaming thresholds are calibrated to flag at most 5\% or 1\% of 1,703 safe validation
responses. We measure unsafe-response detection, exact hits (first flag within the unsafe span), median detection delay from unsafe onset, and false alarms on held-out safe BeaverTails responses (Safe BT). We report mean and standard deviation over three seeds for \method; baseline results use one released checkpoint. Bold indicates the best result in each table column.


\parhead{Baselines}
We use the released Qwen3Guard-Stream-0.6B and SingProbe checkpoints without retraining, as their training data are unavailable. For F1 evaluation, each retains its published decision rule. SingProbe flags responses when any token reaches probability 0.5 and uses its query protocol for prompts. Qwen3Guard-Stream operates in strict mode, requiring two consecutive unsafe tokens. Our evaluation reproduces published F1 within 0.6 points on 12/14 Qwen3Guard-Stream entries and 0.9 points on 7/9 SingProbe entries.


\parhead{Evaluation paths}
Tables~\ref{tab:main}--\ref{tab:stream} use the deployed decode path, with pre-filled responses processed one token at a time through compiled, CUDA-graph-captured dual-row steps. Ablations use faster offline scoring on fixed token sequences, which agrees with deployed
inference (Section~\ref{sec:efficiency}).


\parhead{Latency measurement}
We benchmark on an RTX A6000 (GA102) at batch size one using bf16, compilation, and CUDA graphs. We report the median of 100 decoding steps at 512, 2k, and 8k cached positions, with no competing GPU workloads.\footnote{8k denotes 8,191 cached positions, the maximum supported by Qwen3Guard-Stream before the next token.} All methods use identical generator kernels: cuBLAS below layer $L-K$ and our kernels above it. Qwen3Guard-Stream's guard uses the same kernel configuration. Unless specified otherwise, latency overhead is measured relative to the generator alone.

\section{Results}
\label{sec:results}


Table~\ref{tab:main} summarizes classification accuracy, streaming detection, and inference overhead. \method achieves the highest average F1 while maintaining guard-level detection at nearly the cost of SingProbe.

\begin{table*}[t]
\centering
\caption{Unsafe-class F1 per benchmark, with the average F1 and AUPRC.}
\label{tab:bench}
\footnotesize
\setlength{\tabcolsep}{2.2pt}
\begin{tabular}{@{}lccccccccccc@{}}
\toprule
Response & WildGuard & Aegis 2.0 & BeaverTails & SafeRLHF & HarmBench & Think & XSTest & ExpGuard & & Avg. F1 & Avg. AUPRC \\
\midrule
Qwen3Guard-Stream & \best{76.8} & 81.3 & 84.5 & 62.7 & 82.7 & \best{81.7} & 83.8 & \best{90.1} & & 80.4 & 89.0 \\
SingProbe & 75.8 & 79.0 & 82.8 & 55.4$^\ddagger$ & 83.9$^\ddagger$ & 74.3 & 81.9 & 87.7 & & 77.6 & 86.9 \\
\method (ours) & 76.3\std{1.5} & \best{82.3}\std{0.8} & \best{85.6}\std{0.0} & \best{64.2}\std{1.8} &
\best{84.5}\std{0.6} & 79.8\std{1.4} & \best{91.8}\std{1.5} & 86.9\std{1.0} & & \best{81.4}\std{0.2} &
\best{89.9}\std{0.1} \\
\midrule
Prompt & WildGuard & Aegis 2.0 & ToxicChat & OpenAI Mod. & Aegis 1.0 & SimpleST & HarmBench & XSTest & ExpGuard & Avg. F1
& Avg. AUPRC \\
\midrule
Qwen3Guard-Stream & 87.6 & \best{85.9} & 71.4 & 68.0 & 85.4 & \best{98.0} & 97.2 & 85.3 & 82.2 & 84.6 & 91.7 \\
SingProbe & 86.6 & 84.8 & 71.0 & 67.2 & \best{87.0} & 97.4 & 79.3$^\ddagger$ & 86.5 & 86.9 & 83.0 & 90.8 \\
\method (ours) & \best{87.9}\std{0.7} & 85.0\std{0.2} & \best{72.3}\std{2.7} & \best{68.3}\std{1.6} & 86.6\std{2.2} &
96.5\std{0.8} & \best{98.9}\std{1.1} & \best{88.5}\std{1.9} & \best{93.0}\std{0.6} & \best{86.3}\std{0.4} &
\best{92.0}\std{0.2} \\
\bottomrule
\end{tabular}
\end{table*}



\subsection{Classification Accuracy}
As shown in Table~\ref{tab:bench}, \method achieves the highest F1 among all three methods on 5 of 8 response and 6 of 9 prompt benchmarks. Compared with Qwen3Guard-Stream, it performs better on 12 of 17 benchmarks, with average improvements of 1.0 and 1.7 F1 points for responses and prompts, respectively. The largest gains occur on XSTest responses (+8.0) and ExpGuard prompts (+10.8). \method also improves average F1 on benchmarks absent from its training sources, by 2.7 points for responses and 0.9 for prompts. Average AUPRC follows
the same ranking as F1, suggesting that the improvements are not limited to the selected classification thresholds.

Performance varies across datasets. On ExpGuard responses, \method trails Qwen3Guard-Stream by 3.2 F1 points despite similar AUPRC, indicating sensitivity to threshold selection. Entries marked $^\ddagger$ for SingProbe use dataset versions that differ from those reported in its original paper.

\subsection{Streaming Detection}
\label{sec:streaming}

\begin{table}[t]
\centering
\caption{Token-level streaming detection on Qwen3GuardTest.}
\label{tab:stream}
\footnotesize
\setlength{\tabcolsep}{4pt}
\begin{tabular}{@{}llccc@{}}
\toprule
Budget & Metric & \begin{tabular}[b]{@{}c@{}}Qwen3Guard-\\Stream\end{tabular} & SingProbe & \method (ours) \\
\midrule
\multirow{4}{*}{5\%} & Detected (\%) & \best{84.7} & 71.8 & 83.6\std{0.6} \\
 & Exact hit (\%) & \best{77.1} & 61.5 & 75.6\std{0.1} \\
 & Delay (tokens) & 7 & 7 & \best{5.0} \\
 & Safe BT flagged (\%) & 8.6 & \best{4.8} & 8.3\std{0.7} \\
\midrule
\multirow{4}{*}{1\%} & Detected (\%) & \best{61.0} & 48.5 & 59.5\std{1.6} \\
 & Exact hit (\%) & 43.3 & 37.0 & \best{50.2}\std{2.3} \\
 & Delay (tokens) & 17 & 12 & \best{8.3} \\
 & Safe BT flagged (\%) & 1.9 & \best{1.3} & 1.8\std{0.1} \\
\bottomrule
\end{tabular}
\end{table}




Table~\ref{tab:stream} evaluates streaming detection on 813 unsafe responses from Qwen3GuardTest. At a 5\% false-alarm budget, \method detects 83.6\% of unsafe responses, compared with 84.7\% for Qwen3Guard-Stream and 71.8\% for SingProbe. At the stricter 1\% budget,
\method remains within 1.5 percentage points of the dedicated guard while exceeding SingProbe by 11.0 points.

Beyond detection accuracy, \method identifies unsafe content earlier. Its median detection delay is 5 tokens at the 5\% budget, compared with 7 for both baselines. At 1\%, this delay is 8.3 tokens, versus 17 for Qwen3Guard-Stream and 12 for SingProbe. \method also
achieves a higher exact-hit rate than the guard at 1\% (50.2\% versus 43.3\%). These timing differences may partly reflect the decision rules: Qwen3Guard-Stream requires two consecutive unsafe tokens, whereas \method flags a single threshold crossing.

The teacher's primary benefit appears to be earlier detection. In a one-seed pilot without teacher-derived onsets, \method detected 85.0\% of unsafe responses at the 5\% budget, but with a median delay of 25 tokens. Offline scoring further shows that the median peak safety score rises from 0.002 before the unsafe span to 0.984 after its onset.

On SingStreamBench, SingProbe performs better, reaching 98.6 response-level AUC compared with 92.8 for \method and 94.5 for Qwen3Guard-Stream. This benchmark uses onset annotations generated by the same LLM that annotated SingProbe's unreleased training data. SingProbe also produces fewer false alarms on held-out safe BeaverTails responses.

\subsection{Inference Efficiency}
\label{sec:efficiency}




Across context lengths of 512 to 8k tokens, \method adds only 105--110\us per token, compared with 46--48\us for SingProbe and 2.3--3.5~ms for Qwen3Guard-Stream. Because verification reuses the generator's KV cache, its absolute overhead remains nearly constant as the context grows. Consequently, its relative overhead decreases from 4.0\% to 2.8\% (Table~\ref{tab:main}).

At 2k context, the generator produces 336 tokens/s. This decreases to 325 with \method and 331 with SingProbe, but only 181 with Qwen3Guard-Stream. Moreover, our kernels accelerate the generator itself by 2.0--2.9\% relative to cuBLAS, bringing the complete
\method execution within 0.9\% of the stock cuBLAS generator. This kernel improvement also benefits SingProbe when evaluated on the same generator.

Finally, offline scoring closely matches deployed decoding. Across three seeds, average F1 and detection rates differ by at most 0.1 points, with only 0.16--0.26\% of token-level verdicts changing.

\subsection{Ablation Studies}
\label{sec:abl}

We examine the effects of cache sharing, kernel fusion, adapter configuration, and token-level supervision. Unless stated otherwise, ablations use offline scoring over three seeds.

\parhead{Shared versus independent KV cache}
To isolate the effect of cache sharing, we compare two verifiers that adapt all 28 layers using final-token labels. The shared-cache variant achieves 81.4 response F1 and 86.7 prompt F1, compared with 81.2 and 86.5 for an independently cached verifier. Their AUPRC
values are also comparable. However, an independent cache introduces 9--59\% additional memory traffic per token, versus less than 0.6\% with cache sharing. This comparison covers complete-response classification rather than streaming detection.



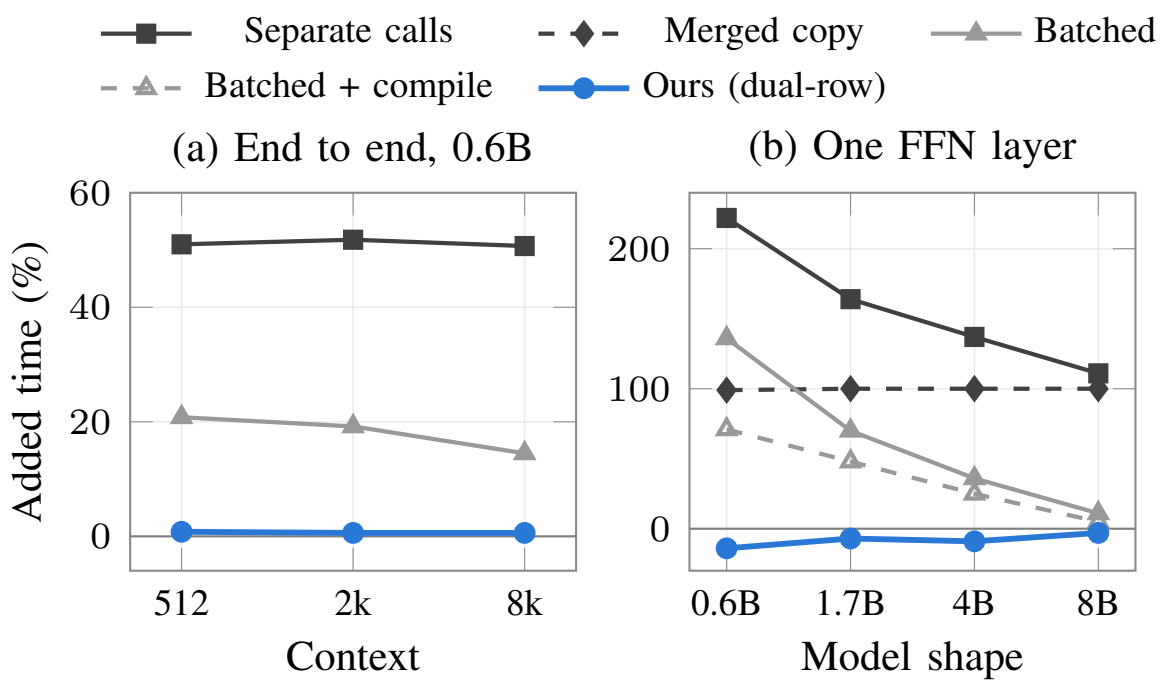
\begin{figure}[t]
\centering
\begin{tikzpicture}[
  sep/.style={black!75, mark=square*},
  mrg/.style={black!75, dashed, mark=diamond*, mark options={solid, scale=1.3}},
  bat/.style={black!40, mark=triangle*, mark options={scale=1.2}},
  batc/.style={black!40, dashed, mark=triangle, mark options={solid, scale=1.2}},
  ours/.style={torablue, mark=*, line width=1.1pt},
]
\begin{groupplot}[
  group style={group size=2 by 1, horizontal sep=0.75cm},
  scale only axis, width=2.95cm, height=2.5cm,
  tick label style={font=\scriptsize}, label style={font=\footnotesize},
  title style={font=\footnotesize, yshift=-5pt},
  xlabel style={yshift=2pt}, ylabel style={yshift=-3pt},
  grid=major, grid style={black!8, line width=0.3pt},
  axis line style={black!45}, tick style={black!45},
  every axis plot/.append style={line width=0.8pt, mark size=1.5pt},
]
\nextgroupplot[title={(a) End to end, 0.6B}, xmin=0.7, xmax=3.3, xtick={1,2,3}, xticklabels={512,2k,8k},
  xlabel={Context}, ymin=-6, ymax=60, ytick={0,20,40,60}, ylabel={Added time (\%)}]
  \draw[black!50, line width=0.4pt] (axis cs:0.7,0) -- (axis cs:3.3,0);
  \addplot[sep] coordinates {(1,51.0) (2,51.8) (3,50.7)};
  \addplot[bat] coordinates {(1,20.8) (2,19.2) (3,14.5)};
  \addplot[ours] coordinates {(1,0.8) (2,0.6) (3,0.6)};
\nextgroupplot[title={(b) One FFN layer}, xmin=0.7, xmax=4.3, xtick={1,2,3,4}, xticklabels={0.6B,1.7B,4B,8B},
  xlabel={Model shape}, ymin=-30, ymax=240, ytick={0,100,200},
  legend to name=leg:exec, legend columns=3,
  legend style={font=\scriptsize, draw=none, /tikz/every even column/.append style={column sep=6pt}}]
  \draw[black!50, line width=0.4pt] (axis cs:0.7,0) -- (axis cs:4.3,0);
  \addplot[sep] coordinates {(1,222) (2,164) (3,137) (4,111)};
  \addlegendentry{Separate calls}
  \addplot[mrg] coordinates {(1,99) (2,100) (3,100) (4,100)};
  \addlegendentry{Merged copy}
  \addplot[bat] coordinates {(1,136) (2,70) (3,36) (4,11)};
  \addlegendentry{Batched}
  \addplot[batc] coordinates {(1,71) (2,48) (3,25) (4,5)};
  \addlegendentry{Batched + compile}
  \addplot[ours] coordinates {(1,-14) (2,-7) (3,-9) (4,-3)};
  \addlegendentry{Ours (dual-row)}
\end{groupplot}
\node[anchor=south] at ($(group c1r1.north west)!0.5!(group c2r1.north east)$) [yshift=9pt]
  {\pgfplotslegendfromname{leg:exec}};
\end{tikzpicture}
\caption{Latency overhead of different execution strategies for the verifier path.}
\label{fig:exec}
\end{figure}


\parhead{Kernel execution strategies}
Fig.~\ref{fig:exec} compares execution strategies against a stock cuBLAS generator. At 0.6B, separate execution adds approximately 51\% latency, while conventional batching adds 15--21\%. Our dual-row kernels reduce this overhead to less than 1\%. For individual FFN layers, merging adapters into separate weight matrices doubles weight traffic and approximately doubles latency. In contrast, our kernels execute both paths 3--14\% faster than cuBLAS executes the generator alone, achieving 69--88\% of peak memory bandwidth.

The advantage persists at larger model sizes. At 1.7B, \method adds 2.8--3.3\% end-to-end latency with an untrained adapter, compared with
14--17\% for conventional batching.


\parhead{Adapter rank and learned features}
Adapter ranks from 4 to 32 produce response F1 scores within 0.9 points of one another (seed 0). To isolate the benefit of feature adaptation, we also train an MLP probe on SingProbe's three hidden-state layers using identical data and final-token labels. The probe achieves 72.5 response F1 and 81.3 prompt F1, compared with 81.5 and 86.7 for \method under the same training conditions. This supports the importance of adapting the generator's upper layers rather than relying exclusively on frozen representations.

\parhead{Adaptation depth and token-level supervision}
With $K=14$, \method achieves 83.5\% streaming detection at the 5\% false-alarm budget, compared with 84.5\% when adapting all 28 layers. The corresponding analytical FLOP overheads at 4k context are 41\% and 84\%, respectively. Token-level supervision is equally important:
increasing $N$ from 0 to 128 improves detection from 19.9\% to 83.5\%, while response F1 remains near 81.5\%.

\section{Limitations}
\label{sec:limitations}

\method targets memory-bound decoding at batch size one; its advantage may diminish under larger batches or speculative decoding~\cite{specdec}. Accuracy is evaluated only on Qwen3-0.6B, with efficiency experiments extending to 1.7B end to end and 8B
for FFN layers. Our cost comparison uses equally sized generator and guard models, and latency measurements cover one GPU. Finally, streaming evaluation replays existing responses rather than generating them online, although the verifier observes the same token sequences.



\section{Conclusion}
\label{sec:conclusion}
We presented \method, a shared-backbone approach that enables continuous safety verification during generation with minimal overhead. By reusing the generator's weights and KV cache through fused dual-row execution, \method achieves detection comparable to a dedicated guard model at nearly the cost of a lightweight probe. Beyond safety monitoring, this architecture offers a foundation for other token-level verification tasks. Supporting multiple concurrent verifiers and evaluating their efficiency remain promising directions for future work.

\newpage
\bibliographystyle{IEEEtran}
\bibliography{references}

\end{document}